\pdfoutput=1
\documentclass[lettersize,journal]{IEEEtran}
\usepackage{graphicx}
\usepackage{tabularx}
\usepackage{multirow}
\usepackage{makecell}  
\usepackage{amsmath,amsfonts,amssymb}
\usepackage{array}
\usepackage{bbding}
\usepackage{booktabs}
\usepackage[table]{xcolor}
\usepackage{textcomp}
\usepackage{arydshln}
\usepackage{stfloats}
\usepackage{pifont}
\usepackage{url}
\usepackage{verbatim}
\usepackage{algorithm}
\usepackage{algpseudocode}
\usepackage{cleveref}

\def\BibTeX{{\rm B\kern-.05em{\sc i\kern-.025em b}\kern-.08em
    T\kern-.1667em\lower.7ex\hbox{E}\kern-.125emX}}
\usepackage{balance}

\newcommand{\xmark}{\text{\ding{55}}}  % pifont
\definecolor{darkgreen}{RGB}{0,127,0}
\definecolor{darkred}{RGB}{200,0,0}
\def\greencheckmark{\textcolor{darkgreen}{\checkmark}}
\def\redxmark{\textcolor{darkred}{\xmark}}

\title{MindHelper: Closed-Loop Embodied Mental-State Reasoning for Precision Intervention}

\author{
Ruoxuan Zhang, Qiaoqiao Wan, Zhengguang Wang, Chenghao Yu, Hongxia Xie\thanks{Corresponding author: Hongxia Xie.}, Wen-Huang Cheng, Jianlong Fu
\thanks{Ruoxuan Zhang,  Qiaoqiao Wan, Zhengguang Wang, Chenghao Yu, and Hongxia Xie are with Jilin University.
 Jianlong Fu is with Microsoft Asia.
 Wen-Huang Cheng is with National Taiwan University.

}
}
\begin{document}

\maketitle
\begin{abstract}
Theory-of-Mind (ToM) reasoning enables embodied agents to understand human beliefs, goals, and intentions, but existing benchmarks mainly evaluate this ability through offline question answering or scenario-level action prediction. MindPower advances embodied ToM by introducing robot-centric reasoning from perception to action; however, it does not evaluate whether an agent can continuously interact with a changing environment and intervene only when assistance is needed. Building on MindPower, we introduce the MindHelper Challenge, which extends embodied ToM evaluation to real-time closed-loop precision intervention. An agent must continuously observe the environment, maintain actor-specific beliefs, identify when a human requires assistance, generate executable actions, and remain silent when intervention is unnecessary. We further propose MindClaw, a simple yet effective Claw-style framework that integrates an actor-specific Belief Table, embodied cognitive skills, and a Trigger-based cognitive dispatcher. Experiments show that MindClaw achieves 36.63\% precise intervention rate and 14.36\% task accuracy, substantially outperforming direct VLM baselines, whose corresponding results remain below 12.05\% and 3.80\%.
\end{abstract}

\section{Introduction}
\label{sec:introduction}

Understanding human mental states is essential for service robots to provide appropriate and timely assistance~\cite{zhang2026social,liu2025modeling}. This ability is closely related to Theory of Mind (ToM)~\cite{leslie2004core,onishi200515,frith2005theory,rao1995bdi,wimmer1983beliefs}, which enables an agent to infer latent mental states, such as beliefs, desires, and intentions, and use them to interpret human behavior. Recent studies~\cite{wimmer1983beliefs,shi2025muma,jin2024mmtom,fan2025somi} have made substantial progress in ToM reasoning. However, most existing methods adopt a role-centric perspective: they infer mental states from the viewpoint of a human character, but do not consider how a service robot should use these states to make decisions and provide assistance. Moreover, these methods are commonly evaluated in a static question-answering setting, where the model only answers questions about human mental states. Consequently, they remain limited to passive mental-state understanding and cannot support active assistance in embodied environments.

\begin{figure}
    \centering
    \includegraphics[width=\linewidth]{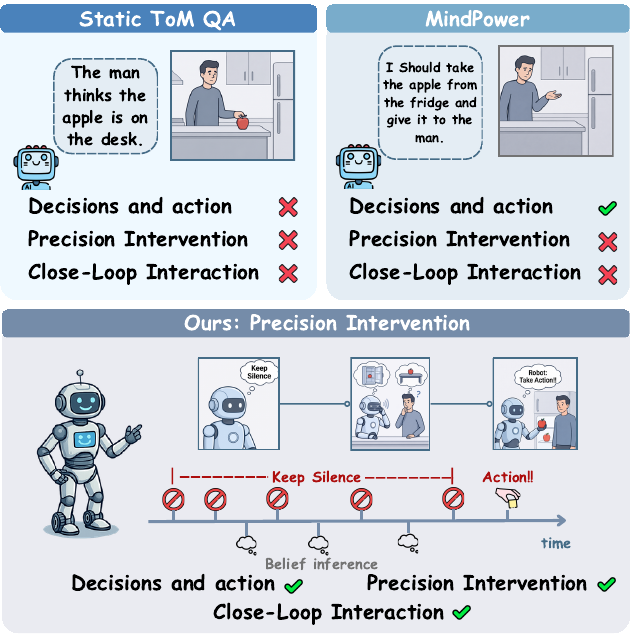}
 \caption{
\textbf{Comparison of static ToM reasoning, MindPower, and MindClaw.}
Static ToM methods only infer human mental states, without producing embodied decisions or actions. MindPower adopts a robot-centric perspective and generates helpful actions based on inferred mental states, but operates only after observing a complete video. In contrast, MindClaw supports closed-loop interaction by continuously observing the environment and tracking human beliefs. It remains silent when assistance is unnecessary and intervenes with an appropriate action when help is needed, enabling timely and precise intervention.
}
    \label{fig:overview}
\end{figure}

MindPower~\cite{zhang2025mindpower} takes an important step toward addressing this limitation by adopting a robot-centric perspective. Instead of only inferring mental states from a human's viewpoint, it requires an embodied agent to infer human beliefs, desires, and intentions and then generate helpful actions. These actions aim to correct false beliefs or complete implicit goals, thereby helping humans accomplish their tasks. However, MindPower still operates in an offline setting: the model observes a complete video and produces its decision and action sequence only after the video ends. It therefore cannot continuously update its understanding or respond while an event is unfolding. As illustrated in Fig.~\ref{fig:overview}, this setting lacks the closed-loop interaction required by service robots operating in dynamic home environments.

A capable service robot should satisfy two key requirements. First, it should support \textbf{Closed-Loop Interaction} by continuously observing the environment, updating its estimate of human mental states, taking actions when needed, and using subsequent observations as feedback. This process allows the robot to provide assistance at the appropriate time. Second, it should achieve \textbf{Precision Intervention}: the robot should intervene only when its assistance is necessary and remain silent when no help is needed. Precision intervention therefore requires the robot to determine both \emph{whether} and \emph{when} to act, rather than producing an action whenever a mental state is inferred.

To bridge this gap, we introduce the \textbf{MindHelper Challenge}, which extends the MindPower benchmark from offline video-based reasoning to online, closed-loop embodied interaction. The challenge evaluates whether current VLM-based agents can continuously observe an evolving home environment, track human mental states, and provide appropriate assistance at the right time. Importantly, an agent must not only decide when and how to help, but also remain silent when intervention is unnecessary. MindHelper therefore evaluates closed-loop precision intervention rather than one-time reasoning over a complete video. Based on this challenge, we further propose \textbf{MindClaw}, a framework that integrates continuous observation, mental-state tracking, decision making, and action execution within a closed loop, enabling embodied agents to achieve timely and precise intervention.

In summary, our main contributions are threefold:
\begin{itemize}
    \item We introduce the \textbf{MindHelper Challenge}, extending MindPower from offline video-based reasoning to online embodied interaction. It establishes two key requirements for service robots: Closed-loop Interaction and Precision Intervention.
    
    \item We propose \textbf{MindClaw}, a closed-loop cognitive control framework for embodied mental-state assistance. It maintains separate environment facts and actor-specific beliefs, and introduces a structured cognitive dispatcher that selects among memory writing, mental-state reasoning, action generation, and silence. The dispatcher combines experience-distilled high-confidence rules with a learned policy, enabling timely intervention without invoking expensive reasoning or action generation at every observation step.
    
    \item Extensive experiments show that MindClaw achieves strong performance across multiple evaluation metrics. Compared with directly applying VLMs at every interaction step, MindClaw reduces unnecessary model calls and token consumption, leading to more efficient and timely assistance.
\end{itemize}

\section{Related Work}
\label{sec:related_work}

\subsection{Theory-of-Mind Benchmarks}
\label{sec:rw_tom_benchmarks}

Recent language and multimodal Theory-of-Mind (ToM) benchmarks have made social and mental-state reasoning measurable in modern foundation models. Text-based benchmarks such as SocialIQA~\cite{sap2019social}, BigToM~\cite{gandhi2023understanding},  FANToM~\cite{kim2023fantom}, Hi-ToM~\cite{wu-etal-2023-hi}, OpenToM~\cite{xu2024opentom}, ToMBench~\cite{chen2024tombench}, and ExploreToM~\cite{sclar2025explore} evaluate whether language models can reason about social situations, nested beliefs, and character-specific mental states in narrative contexts. Multimodal benchmarks such as MMToM-QA~\cite{jin2024mmtom}, BDIQA~\cite{mao2024bdiqa}, CHARTOM~\cite{bharti2024chartom}, MuMA-ToM~\cite{shi2025muma}, MoMentS~\cite{villa2025moments}, SIV-Bench~\cite{kong2026siv}, and MeetingToM~\cite{wang2026meetingtom} further introduce images or videos for reasoning about beliefs, desires, intentions, and other latent social states. Embodied benchmarks such as SoMi-ToM~\cite{fan2025somi} and EnactToM~\cite{juneja2026enacttom} further place ToM evaluation in interactive multi-agent environments.

Nevertheless, most existing benchmarks focus on offline mental-state question answering or task-level multi-agent coordination. They do not jointly evaluate whether an assistive agent can remain connected to a changing environment, maintain actor-specific belief memory, determine when mental-state reasoning should be activated, and decide whether to intervene or remain silent. MindPower~\cite{zhang2025mindpower} moves closer to embodied assistance by introducing a robot-centric reasoning hierarchy from perception to action, but it mainly evaluates final decision and action generation after observing a scenario. MindClaw follows this robot-centric direction while shifting the problem to real-time closed-loop precision intervention, where the agent must decide when to update memory, reason, act, or remain silent.

\subsection{Claw-Based Embodied Interaction Frameworks}
\label{sec:rw_claw_frameworks}

Claw-style agent frameworks connect foundation models with tools, memory, and executable environments through persistent interaction loops. OpenClaw~\cite{openclaw2026} provides a general-purpose runtime for tool invocation, state management, and external execution. Recent extensions adapt this paradigm to embodied systems. RoboClaw~\cite{li2026roboclaw} unifies data collection, policy learning, and long-horizon task execution under a VLM-driven controller. ROSClaw~\cite{cardenas2026rosclaw} connects OpenClaw with ROS~2 through a model-agnostic executive layer that supports capability discovery, action validation, and safe execution. OpenGo~\cite{li2026opengo} enables real-time skill selection for quadruped robots, while StreamingClaw~\cite{chen2026streamingclaw} supports streaming perception, long-term multimodal memory, and a real-time perception--decision--action loop. ABot-Claw~\cite{huo2026abotclawfoundationpersistentcooperative} further extends this architecture toward persistent and cooperative multi-robot execution. Related robotic agent frameworks, such as RoboOS~\cite{tan2025roboos} and AEROS~\cite{qin2026aeros}, similarly organize perception, memory, planning, and reusable robot skills within modular execution architectures. These studies establish a strong foundation for embodied task execution. However, their control states mainly represent the environment, robot capabilities, task progress, and execution history, rather than the perspective-dependent mental states of human actors. Consequently, they do not directly address whether a human’s false belief or implicit goal requires assistance at the current moment. MindClaw complements these frameworks by introducing ToM-driven precision intervention. Its Trigger serves as a cognitive dispatcher that determines when to update an actor-specific belief table, invoke mental-state reasoning, generate an assistance action, or remain silent.

\begin{table*}[t]
\centering
\caption{
Comparison of Theory-of-Mind benchmarks.
``MCQ'' denotes Multiple Choice Question.
``Level-3 Ability'' includes False-Belief Correction,
Implicit Goal Inference \& Completion, and Decision Making and Action.
``Precision Intervention'' evaluates whether an agent intervenes at the
appropriate time and remains silent otherwise.
``Real-Time Interaction'' indicates whether the agent remains connected
to an evolving environment.
}
\label{tab:tom-datasets}

\footnotesize
\renewcommand{\arraystretch}{1.12}
\setlength{\tabcolsep}{2.2pt}
\setlength{\dashlinedash}{1.5pt}
\setlength{\dashlinegap}{1.5pt}

\begin{tabular*}{\textwidth}{
@{\extracolsep{\fill}}
>{\raggedright\arraybackslash}m{0.105\textwidth}
>{\centering\arraybackslash}m{0.075\textwidth}
>{\centering\arraybackslash}m{0.235\textwidth}
>{\centering\arraybackslash}m{0.110\textwidth}
>{\centering\arraybackslash}m{0.075\textwidth}
>{\centering\arraybackslash}m{0.070\textwidth}
>{\centering\arraybackslash}m{0.090\textwidth}
>{\centering\arraybackslash}m{0.085\textwidth}
@{}
}
\toprule
\textbf{Dataset} &
\textbf{Modality} &
\textbf{Output} &
\textbf{Perspective} &
\textbf{Format} &
\makecell{\textbf{Level-3}\\\textbf{Ability}} &
\makecell{\textbf{Precision}\\\textbf{Intervention}} &
\makecell{\textbf{Closed-Loop}\\\textbf{Interaction}} \\
\midrule

Hi-ToM~\cite{wu-etal-2023-hi}
& Text
& Belief
& Role-Centric
& MCQ
& \redxmark
& \redxmark
& \redxmark \\

BigToM~\cite{gandhi2023understanding}
& Text
& Belief, Role's Action
& Role-Centric
& MCQ
& \redxmark
& \redxmark
& \redxmark \\

FANToM~\cite{kim2023fantom}
& Text
& Belief
& Role-Centric
& MCQ
& \redxmark
& \redxmark
& \redxmark \\

MuMA-ToM~\cite{shi2025muma}
& Video, Text
& Belief, Goal
& Role-Centric
& MCQ
& \redxmark
& \redxmark
& \redxmark \\

MMToM-QA~\cite{jin2024mmtom}
& Video, Text
& Belief, Goal
& Role-Centric
& MCQ
& \redxmark
& \redxmark
& \redxmark \\

GridToM~\cite{li2025black}
& Video, Text
& Belief
& Role-Centric
& MCQ
& \redxmark
& \redxmark
& \redxmark \\

SoMi-ToM~\cite{fan2025somi}
& Video, Image
& State, Goal, Behavior
& Role-Centric
& MCQ
& \redxmark
& \redxmark
& \redxmark \\

MindPower~\cite{zhang2025mindpower}
& Video, Text
& Perception, Belief, Desire, Intention, Decision, Action
& Robot-Centric
& Open-Ended
& \greencheckmark
& \redxmark
& \redxmark \\

\hdashline

\textbf{MindClaw}
& \textbf{Video, Text, Real-time input}
& \textbf{Executable Action}
& \textbf{Robot-Centric}
& \textbf{Open-Ended}
& \greencheckmark
& \greencheckmark
& \greencheckmark \\

\bottomrule
\end{tabular*}
\end{table*}
% \begin{figure}
%     \centering
%     \includegraphics[width=\linewidth]{Mindcot.pdf}
%     \caption{MindPower Reasoning Hierarchy. The agent first receives multimodal input, then performs mental reasoning to form beliefs, desires, and intentions, and finally makes decisions and generates action plan based on this reasoning.}
%     \label{fig:mindpower-reasoning}
% \end{figure}

% \begin{figure*}
%     \centering
%     \includegraphics[width=\linewidth]{compare.pdf}
%     \caption{Robot-Centric MindPower Reasoning Hierarchy. Existing benchmarks, such as MuMA-ToM, include only Stage 1 and Stage 2 of the video, and focus solely on inferring the mental reasoning of the human (Alice) in the input video. Our dataset additionally includes Stage 3, where Alice returns to search for the item. Moreover, in Level-2 (Mental Reasoning) of MindPower, we infer the mental reasoning of both the embodied agent and the human, whereas existing ToM Benchmarks only infer the role’s mental state through multiple-choice questions. Detailed example is provided in Sec. B of the Supplementary Material.}
%     \label{fig:compare}
% \end{figure*}

% \bottomrule
% \end{tabular}
% \end{table*}

% \section{Close-Loop Precision Intervention Challenge}

% \section{Challenge}

% \section{Challenge}

\section{MindHelper Challenge}

% \section{MindHelper Challenge}

We introduce the \textbf{MindHelper Challenge}, a benchmark for evaluating closed-loop precision intervention in embodied environments. In MindHelper Challenge, a robot helper remains connected to an evolving environment: it continuously observes environmental changes and human behavior, executes assistance actions, and uses the resulting states for subsequent decisions. The robot must jointly determine whether assistance is needed, when to intervene, and how to assist. It should act only when help is required and remain silent otherwise.

Each environment contains a robot helper, one or more human actors, and a set of interactive objects. We implement the challenge in VirtualHome and ThreeDWorld, both of which provide interactive environments and executable actions. We next describe the task definition and closed-loop interaction setting.

\subsection{Preliminary: MindPower Benchmark}
\label{sec:prelim_mindpower}

MindPower~\cite{zhang2025mindpower} is a Robot-Centric Theory-of-Mind (ToM) benchmark designed to evaluate whether VLM-based embodied agents can link visual perception, mental-state reasoning, decision making, and action generation. As shown in Tab.~\ref{tab:tom-datasets}, it contains 590 video-text examples, where the model takes a video as input and produces structured textual outputs. Unlike benchmarks that only evaluate mental-state inference, MindPower provides labels that include not only belief, desire, and intention reasoning, but also downstream decisions and executable actions. It defines two task types: \emph{False-Belief Correction}, where the agent assists a person who holds an incorrect belief about the environment, and \emph{Implicit Goal Inference and Completion}, where the agent infers an unstated goal from human behavior and completes the missing assistance. MindPower further introduces a reasoning hierarchy that organizes perception, mental reasoning, decision making, and action generation, as shown in Tab.~\ref{tab:tom-datasets}. Compared with prior ToM benchmarks, MindPower focuses on scenarios in which mental reasoning is necessary for selecting appropriate decisions and actions.

% \subsection{Comparison between MindPower and MindHelper}

\subsection{Comparison between MindPower and MindHelper benchmark}

Each MindPower example requires the model to observe a complete scenario and then generate a reasoning hierarchy followed by a final decision or action. This setting supports the evaluation of Robot-Centric ToM reasoning, but remains a static video-to-text formulation. The agent does not interact with a live simulator, maintain actor-specific beliefs over time, or determine whether and when mental reasoning should be activated.

\begin{figure*}
    \centering
    \includegraphics[width=\linewidth]{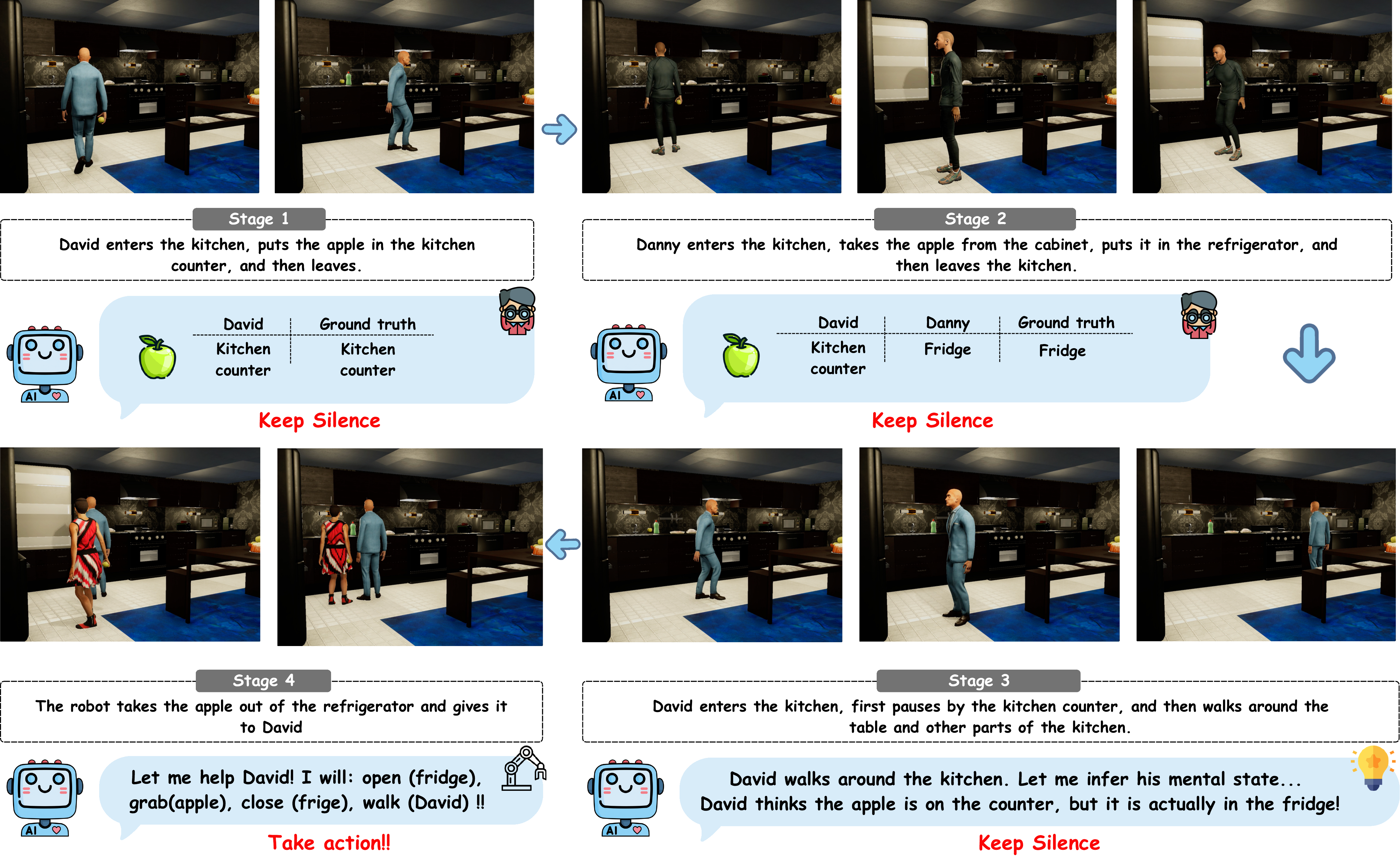}
    \caption{
\textbf{An example from MindHelper Challenge.}
The robot continuously observes the evolving environment and actor states in a closed loop. It remains silent while assistance is unnecessary, even when a false belief exists, and intervenes only when the false belief or implicit goal blocks the actor's task progress. In this example, once David's false belief about the apple's location prevents him from finding it, the robot retrieves the apple from the refrigerator and gives it to him. The red-clothed humanoid avatar represents the robot. Detailed example is shown in the supplementary.
}
    \label{fig:compare}
\end{figure*}

As illustrated in Fig.~\ref{fig:compare}, MindPower takes the complete video covering Stages 1--3 as a single input and produces the final decision and action after observing the entire scenario. The predicted action is not executed in the environment. In MindHelper Challenge, by contrast, the robot agent observes Stages 1--3 sequentially while remaining connected to the simulator. It updates the actor's belief state as the environment changes and remains silent while assistance is unnecessary. Once the false belief blocks the actor's task progress, the agent generates an executable action. The simulator executes this action and returns the resulting state in Stage 4, which closes the observation--action loop.

Existing ToM benchmarks, such as MuMA-ToM~\cite{shi2025muma} and MMToM-QA~\cite{jin2024mmtom}, mainly formulate mental-state understanding as offline question answering. In the example shown in Fig.~\ref{fig:compare}, these benchmarks would observe the evidence provided in the early stages and answer questions about the actor's belief or goal. They do not require the model to determine when to intervene, generate an executable assistance action, or use execution feedback for subsequent decisions.

% MindHelper preserves the Robot-Centric ToM objective of MindPower while extending it to real-time embodied interaction. It introduces actor-specific belief memory, trigger-controlled cognitive operations, and precision intervention, allowing the robot to reason and act only when needed. Table~\ref{tab:tom-datasets} summarizes this broader comparison.

% \begin{table*}[t]
% \centering
% \caption{Key differences between MindPower and MindHelper.}
% \label{tab:mindpower_MindHelper}

% \normalsize
% \renewcommand{\arraystretch}{1.15}
% \setlength{\tabcolsep}{7pt}

% \begin{tabular}{
%     @{}
%     >{\raggedright\arraybackslash}p{0.18\textwidth}
%     >{\raggedright\arraybackslash}p{0.29\textwidth}
%     >{\raggedright\arraybackslash}p{0.41\textwidth}
%     @{}
% }
% \toprule
% \textbf{Dimension} &
% \textbf{MindPower} &
% \textbf{MindHelper} \\
% \midrule

% Input
% & Complete video
% & Continuous windows or interactive streams \\

% Reasoning mode
% & Full reasoning for each sample
% & On-demand triggering \\

% Output
% & Final decision/action
% & Sequence of cognitive operations and final action \\

% Environment
% & Offline video
% & Video, simulator, and user-controlled interaction \\

% Evaluation focus
% & Reasoning and action correctness
% & Task Accuracy (TA) and Precision Intervention Accuracy (PIA) \\

\subsection{Task Setting}

% Each environment contains a robot helper, one or more human actors, and a set of objects. The robot continuously observes changes in the environment and the actors' behavior. Its objective is not to intervene as often as possible, but to determine whether assistance is needed, when to provide it, and how to assist. When assistance is unnecessary, the robot should remain silent.

Following MindPower~\cite{zhang2025mindpower}, we consider two mental-state-driven assistance tasks in MindHelper Challenge:

\begin{itemize}
    \item \textbf{False-Belief Correction:}
    This task evaluates whether the robot can identify an actor's incorrect belief about the environment, such as an outdated belief about an object's location, and provide correction when the belief affects task progress.

    \item \textbf{Implicit Goal Inference \& Completion:}
    This task evaluates whether the robot can infer an unstated goal from behavioral cues, such as searching for an object or repeatedly failing to perform an action, and complete the required task step.
\end{itemize}

The tasks also include scenarios involving actors who may require additional assistance, such as wheelchair users and children.

The assistance process consists of three steps:
\begin{enumerate}
    \item \textbf{Identify the need for assistance.}
    The robot infers whether an actor holds a false belief or has an implicit goal, and determines whether this mental state prevents the actor from completing the task.

    \item \textbf{Ground the assistance target.}
    Based on the actor's belief and goal, the robot identifies the relevant object, action, and location. If the actor is searching for an object, the robot determines its actual location. If the actor cannot complete a required task step, the robot navigates to the relevant location.

    \item \textbf{Execute the assistance action.}
    The robot performs a minimal helpful action, such as retrieving the required object or completing a task step that the actor cannot perform. If intervention is not required, the robot outputs \texttt{none}.
\end{enumerate}
% \subsection{Close-Loop Settings}

% \subsection{Closed-Loop and Precision Intervention Setting}

\subsection{Closed-Loop and Precision Intervention Setting}

At each time step $t$, the robot observes the environment and the actor through an input window $w_t$. Let $e_t$ denote the physical environment state, including the states and locations of objects, and let $h_t$ denote the observable actor state, including the actor's position, action, and interaction with objects. The input window is defined as
\begin{equation}
    w_t = \mathcal{O}(e_t,h_t),
    \label{eq:input_window}
\end{equation}
where $\mathcal{O}$ denotes the observation process. 

Based on $w_t$, the robot jointly determines whether to intervene and how to assist. Its external action is defined as
\begin{equation}
a_t =
\begin{cases}
a_t^\star, & \text{if assistance is required},\\
\mathrm{none}, & \text{otherwise},
\end{cases}
\label{eq:assistance_action}
\end{equation}
where $a_t^\star\in\mathcal{A}_{\mathrm{exec}}$ is an appropriate executable action. A valid intervention must be both helpful and timely~\footnote{More details about the action is in the supplementary.}. If assistance is unnecessary, the robot should output $\mathrm{none}$ to avoid redundant or disruptive actions.

After $a_t$ is executed, the environment and actor states evolve as
\begin{equation}
    (e_{t+1},h_{t+1})
    =
    \mathcal{T}(e_t,h_t,a_t),
    \label{eq:closed_loop_transition}
\end{equation}
where $\mathcal{T}$ denotes the environment transition, including the effect of the robot action and the actor's subsequent behavior. The updated states produce the next input window $w_{t+1}$.

This process forms a closed interaction loop: the robot observes the evolving scene, infers the actor's mental state, intervenes only when needed, and uses the resulting environment state for the next decision. Unlike offline video-to-text reasoning, each robot action can directly affect subsequent observations and task progress.

\subsection{Example}

% \subsection{Example}

Figure~\ref{fig:compare} presents an example of precision intervention. The robot does not intervene simply because the actor holds a false belief. If the false belief does not affect the actor's current activity, the robot remains silent. The robot acts only when the false belief blocks the actor's task progress. This example shows that intervention is determined by the effect of the mental state on the ongoing task, rather than by the presence of a false belief alone.

% \subsection{Comparison with other ToM benchmark}

\section{Method}
\label{sec:method}
\begin{figure*}[t]
    \centering
    \includegraphics[width=0.9\linewidth]{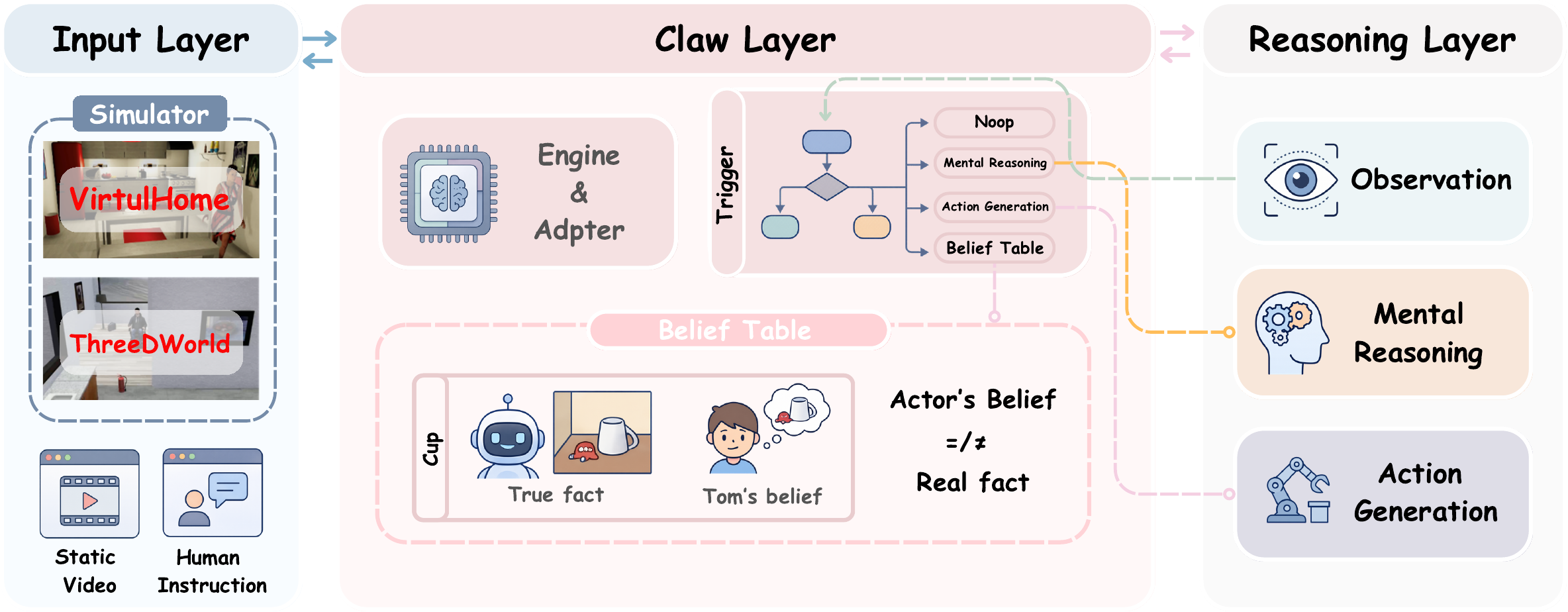}
    \caption{\textbf{Overview of the MindClaw framework.} The Input Layer receives real-time simulator inputs, static videos, and human instructions, while the Claw Layer maintains a belief table that separates real-world facts from actor-specific beliefs. Based on the current observation and memory, the Trigger selects the next operation—belief update, mental reasoning, action generation, or \texttt{noop}.}
    \label{fig:method}
\end{figure*}

% Building on the MindClaw Challenge, we propose the \textbf{MindClaw framework}. The challenge places two key requirements:

% \begin{enumerate}
%     \item \textbf{Timely intervention.}
%     The robot must continuously observe the evolving environment and determine whether and when intervention is necessary.

%     \item \textbf{Memory-grounded assistance.}
%     Accurate intervention requires information accumulated over time, because false beliefs and implicit goals often depend on what the actor previously observed or intended. Relying only on the latest observation window may lose critical evidence and lead to incorrect actions.
% \end{enumerate}

% To address these requirements, we propose MindClaw, a Claw-based framework that maintains actor-specific belief memory and interaction history, and uses a Trigger to determine when to update memory, perform mental-state reasoning, generate an action, or remain silent.

% To address these requirements, we propose MindClaw, a Claw-based framework that maintains actor-specific belief memory, and uses a Trigger to determine when to update memory, perform mental-state reasoning, generate an action, or remain silent. We will introduce the framework in the below sections.

The MindHelper requires an embodied helper to make two coupled decisions in an evolving environment: it must determine whether assistance is needed and, if so, produce an executable action at the appropriate time. This is difficult for a one-shot vision--language policy because the evidence needed for intervention is distributed over time. In particular, the robot must preserve what is true in the environment, what each actor has observed, and whether an inferred mental state currently obstructs the actor's task.

To address this problem, we propose \textbf{MindClaw} (Figure~\ref{fig:method}), a closed-loop cognitive control framework that separates external interaction from internal reasoning. Rather than mapping every observation window directly to a robot action, MindClaw first selects an internal cognitive operation: updating environment facts, updating an actor-specific belief, invoking mental-state reasoning, invoking action generation, or terminating the current cycle without action. This structured interface has two advantages. First, it preserves task-relevant evidence across observation windows. Second, it allows expensive reasoning and action generation to be invoked only when their prerequisites are satisfied.

\subsection{Architecture Overview}
\label{sec:method_overview}

\subsubsection{Input Layer}
% \label{sec:multi_input_interface}

The Input Layer enables MindClaw to process inputs from three types of sources: real-time simulators, including VirtualHome~\cite{puig2018virtualhome} and ThreeDWorld~\cite{gan2020threedworld}, static videos, and human input. VirtualHome and ThreeDWorld serve as executable simulator environments. They provide real-time visual observations and symbolic scene information, while also receiving executable commands from MindClaw. Static video provides offline visual observations for replay and evaluation. Human input, such as keyboard control or user instructions, provides interactive events that may update the task state or influence actor behavior. Together, these input sources allow MindClaw to operate in both interactive environments and offline evaluation settings.

% All input sources are converted into a common window representation. A window may contain rendered frames, simulator state, object and actor identifiers, user-control metadata, task context, and candidate actor/object/location sets. This common representation allows the rest of the framework to remain unchanged across offline video analysis, real-time simulator execution, and human-controlled interaction. In the simulator setting, the interface is closed loop: after MindClaw outputs an action $a_t$, the action is sent to the environment, and the next input window reflects the updated state.

\subsubsection{Claw Layer}
\label{sec:claw_layer}

The Claw Layer serves as the control layer of MindClaw. It maintains the online interaction state and determines when the Reasoning Layer should be called. The Claw Layer consists of four components: Engine, Adapter, Memory, and Trigger.

\textbf{Engine.} The Engine manages the runtime loop of MindClaw. At each time step, it receives the current input from the Input Layer, coordinates the Adapter, Memory, and Trigger, and calls the Reasoning Layer when required. It then applies the state updates selected by the Trigger and stores the resulting interaction records. The Engine keeps MindClaw synchronized with the environment, ensuring that each executed action is reflected in the next input and interaction state.

\textbf{Adapter.} The Adapter converts different input sources into a unified input format. It processes VirtualHome states, ThreeDWorld streams, static video frames, and human input, and maps their source-specific actor identifiers, object names, location names, and action commands to the canonical names used by MindClaw. In this way, the Memory, Trigger, and Reasoning Layer can operate on consistent actor, object, and location slots without handling differences between input sources.

\textbf{Memory.} The Memory stores the information required for closed-loop mental-state reasoning. It consists of two parts. The first is the recent operation history, which records previous Trigger outputs, Reasoning Layer calls, action-generation calls, and execution results. The second is the belief table, which explicitly separates visually grounded facts from actor-specific beliefs:
\begin{equation} 
\begin{aligned} 
\texttt{visual\_facts}[object] &\rightarrow location, \\ 
\texttt{actor\_beliefs}[actor][object] &\rightarrow location.
\end{aligned} 
\end{equation}
This separation allows MindClaw to distinguish the actual environment state from each actor's belief about the environment, which is necessary for selecting appropriate assistance.

\textbf{Trigger.} The Trigger is implemented as an embodied cognitive skill that selects the next internal operation $z_t$. It takes the current observation, recent operation history, and belief table as input. Based on this context, it determines whether MindClaw should update the belief table, perform mental-state reasoning, generate an assistance action, or take no further operation. The Trigger therefore controls the execution flow between the Claw Layer and the Reasoning Layer. Its detailed formulation, implementation, and optimization are introduced in Sec.~\ref{sec:trigger_dispatcher}.

\subsubsection{Reasoning Layer}
\label{sec:reasoning_layer}

The Reasoning Layer contains the model-based components that interpret observations, infer mental states, and generate actions. It has three modules: Observation, Mental Reasoning, and Action Generation.

\textbf{Observation.} The Observation module converts each input window into a structured observation $o_t$. It summarizes visibility changes and task-relevant scene information. The module is descriptive rather than prescriptive: it records what happens in the environment but does not determine whether MindClaw should intervene.

\textbf{Mental Reasoning.} The Mental Reasoning module is called when the Trigger emits \texttt{reasoning\_run}. Given the current observation and belief table, it determines whether the focus actor has a false belief or an implicit goal. It then evaluates whether the identified mental state prevents the actor from completing the task and whether assistance is required.

\textbf{Action Generation.} The Action Generation module is called when the Trigger emits \texttt{action\_run}. It converts the reasoning record $r_t$ into the final robot action $a_t$. If intervention is not required, it outputs $a_t=\texttt{none}$. Otherwise, it generates a minimal helpful action or action sequence, such as removing an obstacle or helping the actor locate an object. In simulator environments, the generated action is converted into executable environment commands and sent to the environment through the Engine.

\begin{figure*}[t]
    \centering
    \includegraphics[width=1.0\linewidth]{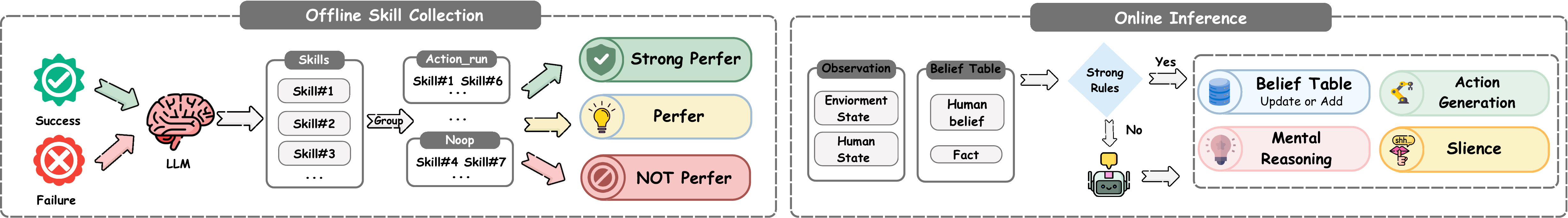}
    \caption{\textbf{Offline cognitive skill collection and online skill-guided Trigger inference.} Successful and failed rollouts are summarized into operation-specific skills. During inference, high-confidence skills are applied as deterministic matchers, while the remaining cases are handled by the skill-conditioned Trigger.}
    \label{fig:skill}
\end{figure*}

\subsection{Trigger as a Cognitive Dispatcher}
\label{sec:trigger_dispatcher}

The Trigger is the cognitive dispatcher of MindClaw. Rather than directly producing a robot action, it selects the next internal cognitive operation based on the structured observation, belief table, recent operation history, and valid grounding candidates. We formulate it as an embodied cognitive policy: its input is grounded in the current environment and actor-specific belief memory, while its output controls internal cognition rather than physical motion. The Trigger therefore controls the execution flow among belief update, Mental Reasoning, Action Generation, and silence in the closed loop.

Let
\begin{equation}
x_t=(o_t,B_{t-1},H_{t-1},C_t)
\label{eq:trigger_context}
\end{equation}
denote the Trigger context, where $o_t$ is the structured observation, $B_{t-1}$ is the current belief table, $H_{t-1}$ is the recent operation history, and $C_t$ contains the valid actor, object, and location candidates. The Trigger outputs a structured cognitive operation
\begin{equation}
z_t=(v_t,i_t,j_t,\ell_t),
\label{eq:trigger_tuple}
\end{equation}
where $v_t$ is the operation verb and $i_t$, $j_t$, and $\ell_t$ are the actor, object, and location slots, respectively. Thus, $z_t$ is an internal control signal rather than the robot's external action.

We model the Trigger as a skill-conditioned policy:
\begin{equation}
\pi_\theta(z_t\mid x_t,\mathcal S)
=P_\theta(v_t,i_t,j_t,\ell_t\mid x_t,\mathcal S),
\label{eq:trigger_policy}
\end{equation}
where $\mathcal S$ is the embodied cognitive skill collection introduced in Sec.~\ref{sec:skill_collection}. Instead of directly mapping an observation to a robot action, this policy first predicts a cognitive operation that updates memory, invokes a reasoning module, or terminates the current decision with silence.

We further factorize the prediction into operation selection and slot grounding:
\begin{equation}
\begin{aligned}
\pi_\theta(z_t\mid x_t,\mathcal S)
&=P_\theta(v_t\mid x_t,\mathcal S)\\
&\quad\cdot P_\theta(i_t,j_t,\ell_t\mid v_t,x_t,\mathcal S).
\end{aligned}
\label{eq:trigger_factorization}
\end{equation}
The first term selects the required cognitive operation, while the second grounds it to valid entities in the current environment. This factorization separates operation-selection errors from slot-grounding errors.

The operation space is
\begin{equation}
\begin{aligned}
\mathcal Z=\{&
\texttt{belief\_create\_visual\_fact}(\texttt{none},j,\ell),\\
&\texttt{belief\_update\_visual\_fact}(\texttt{none},j,\ell),\\
&\texttt{belief\_create\_actor\_belief}(i,j,\ell),\\
&\texttt{belief\_update\_actor\_belief}(i,j,\ell),\\
&\texttt{reasoning\_run}(i,j,\texttt{none}),\\
&\texttt{action\_run}(i,j,\texttt{none}),\\
&\texttt{noop}(\texttt{none},\texttt{none},\texttt{none})\}.
\end{aligned}
\label{eq:operation_space}
\end{equation}
All non-empty slots must be selected from $C_t$, making Trigger prediction a constrained operation-selection and slot-grounding problem.

The Trigger follows a causal execution order. The Observation module first produces $o_t$; required belief-table operations are then performed before \texttt{reasoning\_run}; and Mental Reasoning determines whether the inferred mental state blocks the actor's goal before \texttt{action\_run} is selected. A belief mismatch alone therefore does not cause intervention. The Trigger may instead select \texttt{noop}, making silence an explicit decision when no further cognitive operation or external action is required. Within one observation window, the Trigger can be called repeatedly until it selects \texttt{action\_run} or \texttt{noop}.

\subsection{Offline Skill Collection and Online Skill-Guided Inference}
\label{sec:skill_collection}

As shown in Fig.~\ref{fig:skill}, our skill-guided Trigger consists of two stages: \emph{offline skill collection} and \emph{online skill-guided inference}. The offline stage extracts reusable cognitive skills from successful and failed Trigger rollouts. The online stage keeps the collected skills fixed and uses them to determine which cognitive operation should be executed at each interaction step.

\noindent\textbf{Offline Skill Collection.}
We construct the embodied cognitive skill collection $\mathcal S$ from successful and failed Trigger rollouts in the skill-construction split. Each rollout records the observation, Belief Table, predicted operation, and target operation. A fixed skill summarization model $\mathcal M_{\mathrm{sum}}$ contrasts successful and failed cases and extracts reusable decision conditions, such as when an actor-specific belief should be created, a visual fact should be updated, mental-state reasoning should be invoked, or \texttt{noop} should be selected.

After removing duplicated or semantically equivalent skills, we associate each candidate skill with an operation verb $v\in\mathcal V$ and validate it on the skill-construction trajectories. The validated skills are organized as
\begin{equation}
\mathcal S_v=
\left(
\mathcal S_v^{+},
\mathcal S_v^{\pm},
\mathcal S_v^{-}
\right),
\label{eq:skill_categories}
\end{equation}
where $\mathcal S_v^{+}$ contains strongly preferred conditions, $\mathcal S_v^{\pm}$ contains preferred but non-decisive conditions, and $\mathcal S_v^{-}$ contains negative conditions. A strongly preferred condition provides sufficient evidence for selecting $v$, a preferred condition supports $v$ but still requires model judgment, and a negative condition prevents $v$ from being selected. Whenever possible, strongly preferred and negative conditions are compiled into deterministic structured-field or regular-expression matchers. No validation or test trajectory is used during skill construction.

\begin{algorithm}[t]
\caption{Offline Embodied Cognitive Skill Collection}
\label{alg:skill_collection}
\begin{algorithmic}[1]
\Require Skill-construction rollouts $\mathcal D$
\Require Skill summarization model $\mathcal M_{\mathrm{sum}}$
\Ensure Cognitive skill collection $\mathcal S$

\State Initialize candidate skills $\widetilde{\mathcal S}\gets\emptyset$
\ForAll{paired successful and failed rollouts $(\tau^+,\tau^-)\in\mathcal D$}
\State $\widetilde{\mathcal S}_{\mathrm{pair}}
\gets\mathcal M_{\mathrm{sum}}(\tau^+,\tau^-)$
\State $\widetilde{\mathcal S}
\gets\widetilde{\mathcal S}\cup\widetilde{\mathcal S}_{\mathrm{pair}}$
\EndFor
\State Remove duplicated and semantically equivalent skills
\ForAll{candidate skills $s\in\widetilde{\mathcal S}$}
\State $v\gets\Call{AssociatedOperation}{s}$
\State Validate and categorize $s$ on $\mathcal D$
\State Add $s$ to $\mathcal S_v^c$, where $c\in\{+,\pm,-\}$
\EndFor
\ForAll{$s\in\mathcal S_v^+$ that can be expressed by structured conditions}
\State Compile $s$ into a deterministic matcher $R_s$
\EndFor
\State \Return $\mathcal S$
\end{algorithmic}
\end{algorithm}

\noindent\textbf{Online Skill-Guided Inference.}
During online interaction, MindClaw receives the current observation and the existing Belief Table and constructs the Trigger context $x_t$. The skill collection $\mathcal S$ remains fixed throughout inference. The Trigger first checks whether a strongly preferred condition matches the current context. If a strong skill matches and is not blocked by a negative skill, its deterministic matcher directly returns the corresponding grounded operation. Otherwise, the Trigger uses the current context together with the relevant preferred and negative skills to predict the operation:
\begin{equation}
z_t=
\begin{cases}
R_s(x_t), & \text{if a positive match is found},\\
\pi_\theta(x_t,\mathcal S), & \text{otherwise}.
\end{cases}
\label{eq:skill_inference}
\end{equation}
The selected operation then dispatches MindClaw to update the Belief Table, invoke mental-state reasoning, generate an assistance action, or remain silent through \texttt{noop}. This online procedure combines the precision of deterministic rules with the flexibility of model prediction: high-confidence cases are handled by strong rules, while ambiguous cases are resolved by the skill-conditioned Trigger. Therefore, the offline stage converts rollout experience into reusable cognitive skills, and the online stage applies these skills to support closed-loop precision intervention.

% % \State $H_t \gets H_{t-1} \cup \{z_t\}$
% % \State \Return $z_t$
% % \end{algorithmic}
% % \end{algorithm}

% \section{Data Construction}

\section{Experiments}
\label{sec:experiments}

% \subsection{Setup and Metrics}

\subsection{Setup and Metrics}

\noindent\textbf{Experimental Setup.}
We compare MindClaw with three groups of baselines. First, we include closed-source multimodal models, including GPT-5.4~\cite{openai2026gpt54} and Gemini 3.1 Pro~\cite{gemini2026gemini31pro}. Second, we evaluate open-source vision-language models, including Qwen3-VL at different model scales~\cite{Qwen3-VL} and InternVL3.5-8B~\cite{wang2025internvl3_5}. Third, we consider models designed for video understanding and reasoning, including Video-R1~\cite{feng2025video}, OneThinker~\cite{feng2025onethinker}, and VideoAuto-R1~\cite{liu2026videoautor1}. 
For evaluation, each model is connected to the simulator and continuously observes the ongoing activities in real time. Based on the current observation, the model determines whether intervention is needed and generates an action when necessary. The generated action is then executed in the simulator. The prompts are provided in the supplementary material.

\noindent\textbf{Evaluation Metrics.}
We evaluate the models along two complementary dimensions:
action execution and precision intervention.

\smallskip
\noindent\textit{Action Execution.}
We evaluate action execution using TA, SR, and AC. TA measures task-level completion, while SR and AC are adopted from
MindPower~\cite{zhang2025mindpower} to evaluate the predicted action sequence
at the sequence and atomic-action levels, respectively.
\begin{itemize}
    \item \textbf{Task Accuracy (TA):}
    Measures whether the robot successfully completes the task after
    executing its predicted actions in the simulator. We manually
    inspect the resulting execution videos and report the percentage
    of successfully completed episodes.

    \item \textbf{Success Rate (SR):}
    Following MindPower, SR measures the overall similarity between
    the predicted and ground-truth action sequences:
    \begin{equation}
        \mathrm{SR}
        =
        \frac{2R_1+3R_2+5R_L}{10},
    \end{equation}
    where $R_1$, $R_2$, and $R_L$ denote ROUGE-1, ROUGE-2, and
    ROUGE-L, respectively.

    \item \textbf{Action Correctness (AC):}
    Measures the proportion of ground-truth atomic actions, represented
    as \texttt{action(object)}, that are correctly predicted:
    \begin{equation}
        \mathrm{AC}
        =
        \frac{
        |\mathcal{A}_{\mathrm{pred}}
        \cap
        \mathcal{A}_{\mathrm{gt}}|
        }{
        |\mathcal{A}_{\mathrm{gt}}|
        }.
    \end{equation}
\end{itemize}

\smallskip
\noindent\textit{Precision Intervention.}
We further evaluate whether the robot intervenes only when assistance
is needed and remains silent otherwise.
\begin{itemize}
    \item \textbf{Precise Intervention Rate (PIR):}
Measures the proportion of intervention-required episodes in which the
robot triggers an intervention within the valid intervention window.

\item \textbf{Precise Silence Rate (PSR):}
Measures the proportion of intervention-unnecessary episodes in which
the robot correctly remains silent throughout the interaction.

\item \textbf{Precision Intervention Accuracy (PIA):}
Balances the model's ability to intervene and remain silent:
\begin{equation}
    \mathrm{PIA}
    =
    \frac{\mathrm{PIR}+\mathrm{PSR}}{2}.
\end{equation}
\end{itemize}

\subsection{Results and Analysis}

\begin{table*}[t]
\centering
\caption{
Comparison with existing models on action execution and precision
intervention. SR and AC follow the
evaluation protocol of MindPower~\cite{zhang2025mindpower}.
PIR denotes the precise intervention rate, PSR denotes the silence
correctness rate, and
$\mathrm{Avg.}=(\mathrm{PIR}+\mathrm{PSR})/2$.
Higher values are better for all metrics.
}
\label{tab:intervention_results}

\normalsize
\renewcommand{\arraystretch}{1.12}
\setlength{\tabcolsep}{8pt}

\begin{tabular}{lcccccc}
\toprule
\multirow{2}{*}{\textbf{Model}}
&
\multicolumn{3}{c}{\textbf{Action Evaluation}}
&
\multicolumn{3}{c}{\textbf{Precision Intervention}}
\\
\cmidrule(lr){2-4}
\cmidrule(lr){5-7}
&
\textbf{TA} $\uparrow$
&
\textbf{SR} $\uparrow$
&
\textbf{AC} $\uparrow$
&
\textbf{PIR} $\uparrow$
&
\textbf{PSR} $\uparrow$
&
\textbf{Avg. (PIA)} $\uparrow$
\\
\midrule

\multicolumn{7}{l}{\textit{Closed-Source Multimodal Models}}\\
GPT-5.4~\cite{openai2026gpt54}
    & 3.80
    & 2.83
    & 3.29
    & 10.59
    & 99.51
    & 55.05 \\

Gemini 3.1 Pro~\cite{gemini2026gemini31pro}
    & 2.25
    & 0.00
    & 0.00
    & 0.00
    & 86.61
    & 43.31 \\

\addlinespace
\multicolumn{7}{l}{\textit{Open-Source Multimodal Models}}\\
Qwen3-VL-30B~\cite{Qwen3-VL}
    & 0.00
    & 0.47
    & 0.68
    & 12.05
    & 94.27
    & 53.16 \\

Qwen3-VL-8B~\cite{Qwen3-VL}
    & 0.00
    & 0.00
    & 0.00
    & 0.00
    & 99.96
    & 49.98 \\

Qwen3-VL-4B~\cite{Qwen3-VL}
    & 0.00
    & 0.15
    & 0.21
    & 2.30
    & 99.89
    & 51.10 \\

InternVL3.5-8B~\cite{wang2025internvl3_5}
    & 0.00
    & 0.00
    & 0.00
    & 1.14
    & 99.87
    & 50.50 \\

\addlinespace
\multicolumn{7}{l}{\textit{Video Reasoning Models}}\\
Video-R1~\cite{feng2025video}
    & 0.00
    & 0.18
    & 0.25
    & 12.00
    & 99.31
    & 5.65 \\

OneThinker~\cite{feng2025onethinker}
    & 0.00
    & 0.00
    & 0.00
    & 0.00
    & 100.00
    & 50.00 \\

VideoAuto-R1~\cite{liu2026videoautor1}
    & 0.00
    & 0.15
    & 0.22
    & 7.14
    & 94.80
    & 50.97 \\

\midrule
\multicolumn{7}{l}{\textit{Our Method}}\\
\textbf{MindClaw}
    & \textbf{14.36}
    & \textbf{4.12}
    & \textbf{5.88}
    & 35.71
    & 99.90
    & \textbf{67.81} \\

\bottomrule
\end{tabular}
\end{table*}

\begin{figure}
    \centering
    \includegraphics[width=1.0\linewidth]{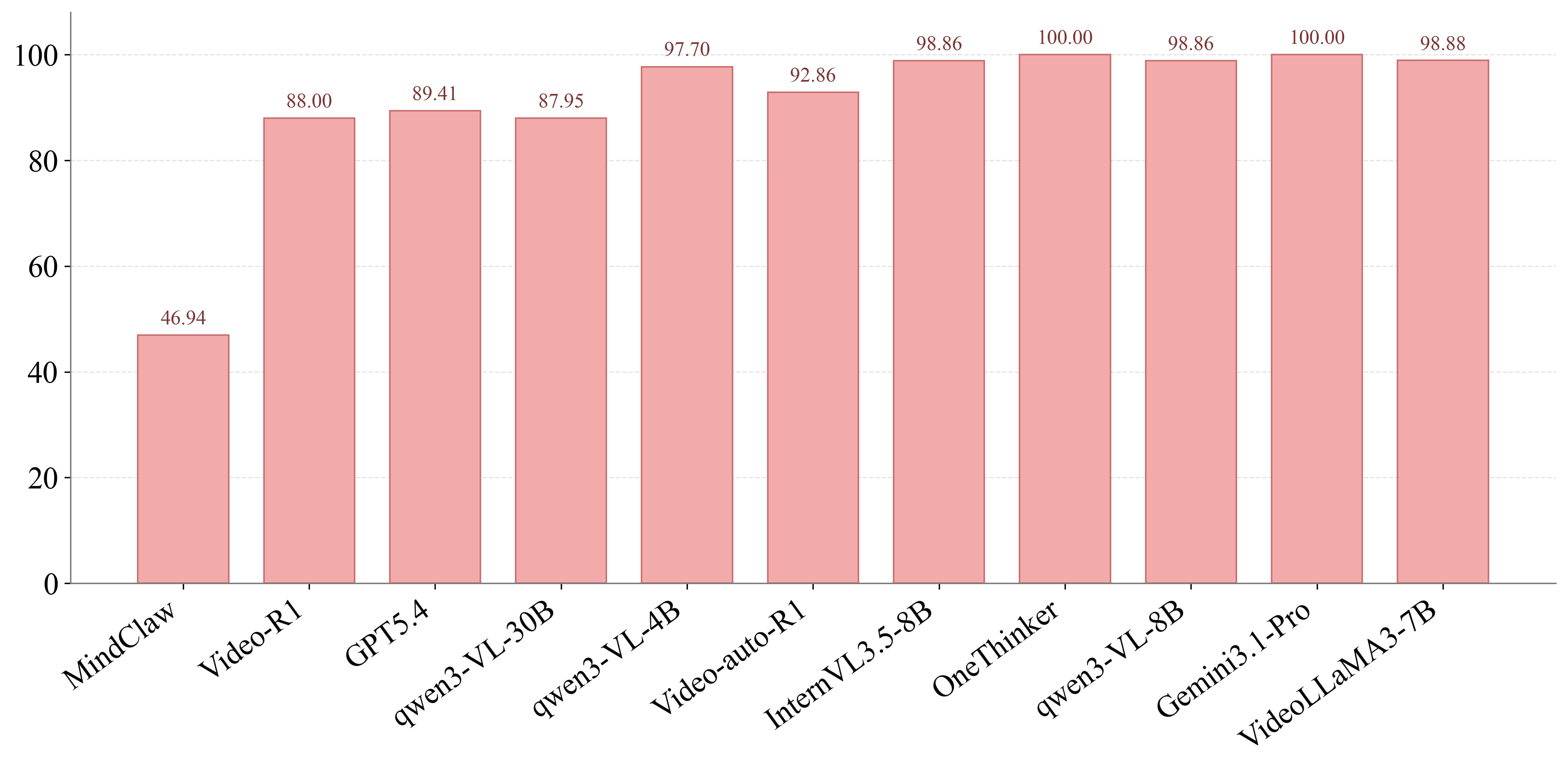}
    \caption{\textbf{Proportion of episodes in which each model remains silent throughout the entire interaction.} The statistic is computed over the full test set, including samples where intervention is required and those where it is unnecessary. The high silence rates reveal that many VLMs tend to avoid intervention rather than accurately determining when silence is appropriate. Lower is better.}
    \label{fig:mir}
\end{figure}

\noindent\textbf{Can existing VLMs perform effective precision intervention?}
As shown in Table~\ref{tab:intervention_results}, existing VLMs perform poorly on closed-loop precision intervention. Although most baselines achieve high PSR, ranging from 86.61\% to 100.00\%, their PIR remains below 12.05\%. Their seemingly moderate average scores are therefore mainly driven by a strong tendency to remain silent, rather than by an ability to determine when intervention is necessary. OneThinker and Qwen3-VL-8B are clear examples: both achieve nearly perfect PSR but obtain a PIR of 0.00\%. To determine whether this result reflects correct intervention decisions or a degenerate silence
strategy, we further measure the proportion of episodes in which each model remains silent throughout the entire interaction. This analysis is conducted over the full dataset, including both intervention-required and intervention-unnecessary episodes. As shown in Figure~\ref{fig:mir}, OneThinker remains silent in 100.00\% of the episodes, while Qwen3-VL-8B remains silent in 98.86\%. Their high PSR therefore does not indicate that they correctly recognize when silence is appropriate; instead, it mainly results from their general tendency to avoid intervention.

Existing VLMs also struggle to produce effective actions. Even GPT-5.4, the strongest baseline in terms of TA, achieves only 3.80\%, while most open-source models obtain zero TA. Their low SR and AC further show that the generated action sequences rarely match those required to complete the task. These results reveal two major limitations of existing VLMs: they fail to identify the appropriate moment at which a human needs assistance, and they fail to generate the correct action sequence needed to provide that assistance. Thus, strong visual
understanding or video reasoning alone is insufficient for timely and effective assistance in a closed-loop interactive environment.

\noindent\textbf{How does MindClaw compare with existing models?}
MindClaw consistently outperforms all baselines, achieving 14.36\% TA, 36.63\% PIA, and 100.00\% CS. Compared with the strongest baseline for each metric, MindClaw improves TA by 10.56 \% and PIA by 17.33 \%. This improvement shows that MindClaw can better determine when intervention is required. 

% \noindent\textbf{How does MindClaw compare with existing models?}

% Table~\ref{tab:intervention_results} compares MindClaw with closed-source models, open-source VLMs, and video reasoning models. Overall, all baseline models perform poorly on this task. Even GPT-5.4, the strongest baseline in terms of TA, achieves only 3.80\%. Most open-source models obtain a TA close to zero, showing that generating plausible responses does not necessarily lead to successful task completion in the simulator.
% A similar limitation is observed in intervention timing. The highest baseline PIA is only 19.30\%, achieved by VideoLLaMA3-7B. This result indicates that existing models have difficulty determining when human assistance is actually needed. 

% MindClaw consistently outperforms all baselines, achieving 14.36\% TA and 36.63\% PIA. Compared with the strongest baselines, it improves TA by 10.56\% and PIA by 17.33\%. These results show that MindClaw is better able to identify when intervention is necessary and convert its reasoning into effective actions. Nevertheless, the relatively low TA also shows that closed-loop embodied assistance remains a challenging task.

\subsection{Ablation Study}

\noindent \textbf{Importance of Cognitive Skills.}
As shown in Table~\ref{tab:ablation_components}, removing the embodied cognitive skills reduces PIA from 67.81\% to 65.26\% and, more substantially, reduces TA from 14.36\% to 3.06\% . To understand this degradation, we further analyze the Trigger errors. When an actor-specific belief should be created, the model incorrectly selects \texttt{noop} in 60.55\% of the cases. Consequently, important belief information is not written into memory, preventing the subsequent mental reasoning and action generation modules from being activated at the appropriate time. This problem is particularly severe in false-belief scenarios, where successful intervention depends on tracking the difference between the actor's belief and the actual environment state. These results show that the cognitive skills are essential for coordinating belief updates, mental reasoning, and action generation.

\noindent \textbf{Importance of the Belief Table.}
As shown in Table~\ref{tab:ablation_components}, removing the Belief Table reduces PIA from 67.81\% to 61.39\% and causes TA to drop from 14.36\% to 0.00\%. Without an explicit actor-specific belief representation, the model cannot reliably track what each actor has observed or distinguish an actor's belief from the actual environment state. To understand this failure, we further analyze the Trigger decisions. Without the Belief Table, the Trigger fails to activate the Action Model in 76.09\% of the cases where action generation is required. This result shows that the Belief Table is essential not only for maintaining actor-specific beliefs, but also for enabling the Trigger to recognize when assistance is needed.

\noindent\textbf{Robustness to the Trigger Backbone.}
We further examine whether the Trigger depends on a specific language-model backbone. Under the same cognitive skills and operation space, GPT-5.5 and Qwen3-4B achieve similar operation accuracies of 89.71\% and 88.76\%, respectively. After excluding \texttt{noop} decisions, their accuracies are 64.14\% and 60.54\%. The small performance gap indicates that the Trigger is robust to the choice of backbone and that its effectiveness mainly comes from the structured operation space and cognitive skill guidance.

\begin{table}[t]
\centering
\caption{
Ablation study of the belief table and embodied cognitive skills.
Avg. denotes the balanced average of the precise intervention rate
(PIR) and precise silence rate (PSR).
}
\label{tab:ablation_components}

\renewcommand{\arraystretch}{1.12}
\setlength{\tabcolsep}{12pt}

\begin{tabular}{lcc}
\toprule
\textbf{Variant}
&
\textbf{PIA} $\uparrow$
&
\textbf{TA} $\uparrow$
\\
\midrule
\textbf{MindClaw (Full)}
    & \textbf{67.81}
    & \textbf{14.36} \\

\quad w/o Cognitive Skills
    & 65.26
    & 3.06 \\

\quad w/o Belief Table
    & 61.39
    & 0.00 \\
\bottomrule
\end{tabular}
\end{table}

% \subsection{Qualitative Evaluation}

\subsection{Efficiency Analysis}

\noindent\textbf{ How efficient is MindClaw?}
The Trigger reduces computational cost by invoking each module only when needed. It skips both mental reasoning and action generation for \texttt{noop}, and invokes only the selected module for other operations. This conditional execution avoids unnecessary model calls and token consumption.

As shown in Table~\ref{tab:token_efficiency}, MindClaw consumes an average of 4,511.24 tokens per episode and 155.42 tokens per call. In comparison, GPT-5.4 consumes 9,036.03 tokens per episode, while Qwen3-VL-4B consumes 8,569.65 tokens. MindClaw therefore reduces token consumption per episode by 50.07\% compared with GPT-5.4 and by 47.36\% compared with Qwen3-VL-4B. It also reduces the average tokens per call by 35.84\% and 21.59\%, respectively. These results show that the Trigger improves efficiency by invoking only the necessary model or operation at each step.

\begin{table}[t]
\centering
\caption{Comparison of average token consumption.}
\label{tab:token_efficiency}

\normalsize
\renewcommand{\arraystretch}{1.08}
\setlength{\tabcolsep}{6pt}

\begin{tabular}{lccl}
\hline
Model & Tokens/Episode & Tokens/Call  &TA\\
\hline
GPT-5.4     & 9,036.03          & 242.23  &3.80\\
Qwen3-VL-4B & 8,569.65          & 198.21  &0.00\\
\hline
\textbf{MindClaw}
            & \textbf{4,511.24} & \textbf{155.42}  &\textbf{14.36}\\
\hline
\end{tabular}
\end{table}

\section{Conclusion}
\label{sec:conclusion}

In this work, we extended MindPower from offline, scenario-level ToM reasoning to the MindHelper Challenge, which evaluates real-time closed-loop precision intervention in embodied environments. We further proposed MindClaw, a simple yet effective framework that combines an actor-specific Belief Table, embodied cognitive skills, and a Trigger-based cognitive dispatcher to determine when to update memory, perform mental-state reasoning, generate an assistance action, or remain silent. MindClaw achieves 14.36\% TA, substantially outperforming direct VLM baselines. Its comparable Trigger performance across GPT-5.5 and Qwen3-4B also demonstrates robustness to different model backbones. Moreover, by invoking reasoning and action generation only when needed, MindClaw reduces average token consumption from 8,570--9,036 to 4,511 tokens per episode. A current limitation is that MindHelper mainly inherits the household environments and indoor scenarios of MindPower; therefore, its generalization to more diverse open-world and outdoor environments remains unverified. Future work will extend MindHelper to broader environments, human activities, and real-world robotic interactions.

% We presented MindClaw, a closed-loop framework for real-time interactive mental-state reasoning in embodied environments. Within this framework, the agent can decide whether a person is acting under a false belief, hidden goal, or task-blocking mismatch, and can produce a reasonable helpful action when assistance is needed.
% We also introduced precision intervention as the central objective for embodied mental-state assistance. An embodied agent should not help by default. It should intervene at the right time, with the right target, and only when the current mental-state and environment state indicate that help is useful. To support this objective, we formulated the trigger as an embodied cognitive skill that mediates between observation, belief memory, mental reasoning, and action generation. By collecting and optimizing trigger skills, MindClaw improves the accuracy of intermediate cognitive-operation prediction and achieves the best reported performance in our experiments.

\bibliographystyle{IEEEtran}
\bibliography{refs}

@String(CVPR= {IEEE Conf. Comput. Vis. Pattern Recog.})

@String(AAAI = {AAAI})

@String(CVPR  = {CVPR})

@inproceedings{rao1995bdi,
  title={BDI agents: From theory to practice.},
  author={Rao, Anand S and Georgeff, Michael P and others},
  booktitle={Icmas},
  volume={95},
  pages={312--319},
  year={1995}
}

@inproceedings{puig2018virtualhome,
  title={Virtualhome: Simulating household activities via programs},
  author={Puig, Xavier and Ra, Kevin and Boben, Marko and Li, Jiaman and Wang, Tingwu and Fidler, Sanja and Torralba, Antonio},
  booktitle={Proceedings of the IEEE conference on computer vision and pattern recognition},
  pages={8494--8502},
  year={2018}
}

@article{leslie2004core,
  title={Core mechanisms in ‘theory of mind’},
  author={Leslie, Alan M and Friedman, Ori and German, Tim P},
  journal={Trends in cognitive sciences},
  volume={8},
  number={12},
  pages={528--533},
  year={2004},
  publisher={Elsevier}
}

@article{gan2020threedworld,
  title={ThreeDWorld: A platform for interactive multi-modal physical simulation},
  author={Gan, C and Schwartz, J and Alter, S and Schrimpf, M and Traer, J and De Freitas, J and Kubilius, J and Bhandwaldar, A and Haber, N and Sano, M and others},
  journal={Advances in Neural Information Processing Systems (NeurIPS)},
  year={2021}
}

@inproceedings{kim2023fantom,
  title={FANToM: A Benchmark for Stress-testing Machine Theory of Mind in Interactions},
  author={Kim, Hyunwoo and Sclar, Melanie and Zhou, Xuhui and Bras, Ronan and Kim, Gunhee and Choi, Yejin and Sap, Maarten},
  booktitle={Proceedings of the 2023 Conference on Empirical Methods in Natural Language Processing},
  pages={14397--14413},
  year={2023}
}

@inproceedings{shi2025muma,
  title={Muma-tom: Multi-modal multi-agent theory of mind},
  author={Shi, Haojun and Ye, Suyu and Fang, Xinyu and Jin, Chuanyang and Isik, Leyla and Kuo, Yen-Ling and Shu, Tianmin},
  booktitle={Proceedings of the AAAI Conference on Artificial Intelligence},
  volume={39},
  number={2},
  pages={1510--1519},
  year={2025}
}

@inproceedings{jin2024mmtom,
    title = "{MMT}o{M}-{QA}: Multimodal Theory of Mind Question Answering",
    author = "Jin, Chuanyang  and
      Wu, Yutong  and
      Cao, Jing  and
      Xiang, Jiannan  and
      Kuo, Yen-Ling  and
      Hu, Zhiting  and
      Ullman, Tomer  and
      Torralba, Antonio  and
      Tenenbaum, Joshua  and
      Shu, Tianmin",
    editor = "Ku, Lun-Wei  and
      Martins, Andre  and
      Srikumar, Vivek",
    booktitle = "Proceedings of the 62nd Annual Meeting of the Association for Computational Linguistics (Volume 1: Long Papers)",
    month = aug,
    year = "2024",
    address = "Bangkok, Thailand",
    publisher = "Association for Computational Linguistics",
    url = "https://aclanthology.org/2024.acl-long.851/",
    doi = "10.18653/v1/2024.acl-long.851",
    pages = "16077--16102"
}

@inproceedings{fan2025somi,
title={SoMi-ToM: Evaluating Multi-Perspective Theory of Mind in Embodied Social Interactions},
author={Xianzhe Fan and Xuhui Zhou and Chuanyang Jin and Kolby Nottingham and Hao Zhu and Maarten Sap},
booktitle={The Thirty-ninth Annual Conference on Neural Information Processing Systems Datasets and Benchmarks Track},
year={2025},
url={https://openreview.net/forum?id=7zFLFtqBm0}
}

@inproceedings{li2025black,
title={From Black Boxes to Transparent Minds: Evaluating and Enhancing the Theory of Mind in Multimodal Large Language Models},
author={Xinyang Li and Siqi Liu and Bochao Zou and Jiansheng Chen and Huimin Ma},
booktitle={Forty-second International Conference on Machine Learning},
year={2025},
url={https://openreview.net/forum?id=CDillQjA7N}
}

@inproceedings{wu-etal-2023-hi,
    title = "Hi-{T}o{M}: A Benchmark for Evaluating Higher-Order Theory of Mind Reasoning in Large Language Models",
    author = "Wu, Yufan  and
      He, Yinghui  and
      Jia, Yilin  and
      Mihalcea, Rada  and
      Chen, Yulong  and
      Deng, Naihao",
    editor = "Bouamor, Houda  and
      Pino, Juan  and
      Bali, Kalika",
    booktitle = "Findings of the Association for Computational Linguistics: EMNLP 2023",
    month = dec,
    year = "2023",
    address = "Singapore",
    publisher = "Association for Computational Linguistics",
    url = "https://aclanthology.org/2023.findings-emnlp.717/",
    doi = "10.18653/v1/2023.findings-emnlp.717",
    pages = "10691--10706"
}

@article{gandhi2023understanding,
  title={Understanding social reasoning in language models with language models},
  author={Gandhi, Kanishk and Fr{\"a}nken, Jan-Philipp and Gerstenberg, Tobias and Goodman, Noah},
  journal={Advances in Neural Information Processing Systems},
  volume={36},
  pages={13518--13529},
  year={2023}
}

@inproceedings{mao2024bdiqa,
  title={BDIQA: A new dataset for video question answering to explore cognitive reasoning through theory of mind},
  author={Mao, Yuanyuan and Lin, Xin and Ni, Qin and He, Liang},
  booktitle={Proceedings of the AAAI Conference on Artificial Intelligence},
  volume={38},
  number={1},
  pages={583--591},
  year={2024}
}

@inproceedings{villa2025moments,
    title = "{M}o{M}ent{S}: A Comprehensive Multimodal Benchmark for Theory of Mind",
    author = "Villa-Cueva, Emilio  and
      Ahmed, S M Masrur  and
      Chevi, Rendi  and
      Cruz, Jan Christian Blaise  and
      Elzeky, Kareem  and
      Cristobal, Fermin  and
      Aji, Alham Fikri  and
      Wang, Skyler  and
      Mihalcea, Rada  and
      Solorio, Thamar",
    editor = "Christodoulopoulos, Christos  and
      Chakraborty, Tanmoy  and
      Rose, Carolyn  and
      Peng, Violet",
    booktitle = "Findings of the Association for Computational Linguistics: EMNLP 2025",
    month = nov,
    year = "2025",
    address = "Suzhou, China",
    publisher = "Association for Computational Linguistics",
    url = "https://aclanthology.org/2025.findings-emnlp.1230/",
    pages = "22591--22611",
    ISBN = "979-8-89176-335-7"
}

@misc{openai2026gpt54,
  title        = {GPT-5.4 Thinking System Card},
  author       = {{OpenAI}},
  year         = {2026},
  howpublished = {\url{https://openai.com/index/gpt-5-4-thinking-system-card/}},
  note         = {Accessed: 2026-05-31}
}

@misc{gemini2026gemini31pro,
  title        = {{Gemini 3.1 Pro Model Card}},
  author       = {{Gemini Team}},
  year         = {2026},
  howpublished = {\url{https://deepmind.google/models/model-cards/gemini-3-1-pro/}},
  note         = {Google DeepMind Technical Report. Accessed: 2026-05-31}
}

@article{onishi200515,
  author = {Onishi, Kristine and Baillargeon, Renée},
  year = {2005},
  month = {05},
  pages = {255-8},
  title = {Do 15-Month-Old Infants Understand False Beliefs?},
  volume = {308},
  journal = {Science (New York, N.Y.)},
  doi = {10.1126/science.1107621}

}

@article{frith2005theory,
  title={Theory of mind},
  author={Frith, Chris and Frith, Uta},
  journal={Current biology},
  volume={15},
  number={17},
  pages={R644--R645},
  year={2005},
  publisher={Elsevier}
}

@article{zhang2026social,
  title={Social World Models: A Comprehensive Survey of Theory of Mind and Social Coordination},
  author={ZHANG, RUOXUAN and LIAO, ZIQI and ZHOU, ZHIYU and JIANG-LIN, JIAN-YU and WU, SIYU and MAO, YANBO and WEN, BIN and XIE, HONGXIA and FU, JIANLONG and YAO, MEIBAO and others},
  journal={ACM Comput. Surv},
  volume={1},
  number={1},
  year={2026}
}

@article{chen2026streamingclaw,
  title={Streamingclaw technical report},
  author={Chen, Jiawei and Chen, Zhe and Du, Chaoqun and He, Maokui and He, Wei and Li, Hengtao and Li, Qizhen and Liu, Zide and Ma, Hao and Pan, Xuhao and others},
  journal={arXiv preprint arXiv:2603.22120},
  year={2026}
}

@inproceedings{sclar2025explore,
  title={Explore theory of mind: program-guided adversarial data generation for theory of mind reasoning},
  author={Sclar, Melanie and Dwivedi-Yu, Jane and Fazel-Zarandi, Maryam and Tsvetkov, Yulia and Bisk, Yonatan and Choi, Yejin and Celikyilmaz, Asli},
  booktitle={International Conference on Learning Representations},
  volume={2025},
  pages={67635--67660},
  year={2025}
}

@article{bharti2024chartom,
  title={CHARTOM: A Visual Theory-of-Mind Benchmark for LLMs on Misleading Charts},
  author={Bharti, Shubham and Cheng, Shiyun and Rho, Jihyun and Zhang, Jianrui and Cai, Mu and Lee, Yong Jae and Rau, Martina and Zhu, Xiaojin},
  journal={arXiv preprint arXiv:2408.14419},
  year={2024}
}

@inproceedings{kong2026siv,
  title={Siv-bench: A video benchmark for social interaction understanding and reasoning},
  author={Kong, Fanqi and Zu, Weiqin and Chen, Xinyu and Yang, Yaodong and Zhu, Song-Chun and Feng, Xue},
  booktitle={Findings of the Association for Computational Linguistics: ACL 2026},
  pages={37379--37403},
  year={2026}
}

@article{wang2026meetingtom,
  title={MeetingToM: Evaluating Multimodal LLMs on Theory-of-Mind Reasoning in Multi-Party Meetings},
  author={Wang, Ziyi and Wu, Yuhang and Piao, Dongxu and Liu, Xingyu and Zhou, Tianhui and Liu, Miao},
  journal={arXiv preprint arXiv:2607.19235},
  year={2026}
}

@article{juneja2026enacttom,
  title={EnactToM: An Evolving Benchmark for Functional Theory of Mind in Embodied Agents},
  author={Juneja, Gurusha and Lu, Dylan and Agashe, Saaket and Diwane, Parth and Gunn, Edward and Srinivasa, Jayanth and Liu, Gaowen and Wang, William Yang and Du, Yali and Wang, Xin Eric},
  journal={arXiv preprint arXiv:2605.09826},
  year={2026}
}

@inproceedings{xu2024opentom,
  title={OpenToM: A comprehensive benchmark for evaluating theory-of-mind reasoning capabilities of large language models},
  author={Xu, Hainiu and Zhao, Runcong and Zhu, Lixing and Du, Jinhua and He, Yulan},
  booktitle={Proceedings of the 62nd Annual Meeting of the Association for Computational Linguistics (Volume 1: Long Papers)},
  pages={8593--8623},
  year={2024}
}

@article{liu2025modeling,
  title={Modeling the mental world for embodied AI: A comprehensive review},
  author={Liu, Biyuan and Xu, Daigang and Jiang, Lei and Guo, Wenjun and Chen, Ping},
  journal={arXiv preprint arXiv:2601.02378},
  year={2025}
}

@article{qin2026aeros,
  title={AEROS: A single-agent operating architecture with embodied capability modules},
  author={Qin, Xue and Luan, Simin and See, John and Yang, Cong and Li, Zhijun},
  journal={arXiv preprint arXiv:2604.07039},
  year={2026}
}

@article{tan2025roboos,
  title={Roboos: A hierarchical embodied framework for cross-embodiment and multi-agent collaboration},
  author={Tan, Huajie and Hao, Xiaoshuai and Chi, Cheng and Lin, Minglan and Lyu, Yaoxu and Cao, Mingyu and Liang, Dong and Chen, Zhuo and Lyu, Mengsi and Peng, Cheng and others},
  journal={arXiv preprint arXiv:2505.03673},
  year={2025}
}

@article{cardenas2026rosclaw,
  title={ROSClaw: An OpenClaw ROS 2 framework for agentic robot control and interaction},
  author={Cardenas, Irvin Steve and Arnett, Marcus Anthony and Yeo, Natalie Catherine and Sah, Lucky and Kim, Jong-Hoon},
  journal={arXiv preprint arXiv:2603.26997},
  year={2026}
}

@article{li2026opengo,
  title={OpenGo: An OpenClaw-based robotic dog with real-time skill switching},
  author={Li, Hanbing and Cao, Xuewei and Zeng, Zhiwen and Wu, Yuhan and Zhang, Yanyong and Xia, Yan},
  journal={arXiv preprint arXiv:2604.01708},
  year={2026}
}

@inproceedings{sap2019social,
  title={Social IQa: Commonsense reasoning about social interactions},
  author={Sap, Maarten and Rashkin, Hannah and Chen, Derek and Le Bras, Ronan and Choi, Yejin},
  booktitle={Proceedings of the 2019 conference on empirical methods in natural language processing and the 9th international joint conference on natural language processing (EMNLP-IJCNLP)},
  pages={4463--4473},
  year={2019}
}

@inproceedings{chen2024tombench,
  title={ToMBench: Benchmarking theory of mind in large language models},
  author={Chen, Zhuang and Wu, Jincenzi and Zhou, Jinfeng and Wen, Bosi and Bi, Guanqun and Jiang, Gongyao and Cao, Yaru and Hu, Mengting and Lai, Yunghwei and Xiong, Zexuan and others},
  booktitle={Proceedings of the 62nd Annual Meeting of the Association for Computational Linguistics (Volume 1: Long Papers)},
  pages={15959--15983},
  year={2024}
}

@article{wimmer1983beliefs,
  title={Beliefs about beliefs: Representation and constraining function of wrong beliefs in young children's understanding of deception},
  author={Wimmer, Heinz and Perner, Josef},
  journal={Cognition},
  volume={13},
  number={1},
  pages={103--128},
  year={1983},
  publisher={Elsevier}
}

@article{liu2026videoautor1,
  title={VideoAuto-R1: Video Auto Reasoning via Thinking Once, Answering Twice},
  author={Liu, Shuming and Zhuge, Mingchen and Zhao, Changsheng and Chen, Jun and Wu, Lemeng and Liu, Zechun and Zhu, Chenchen and Cai, Zhipeng and Zhou, Chong and Liu, Haozhe and Chang, Ernie and Suri, Saksham and Xu, Hongyu and Qian, Qi and Wen, Wei and Varadarajan, Balakrishnan and Liu, Zhuang and Xu, Hu and Bordes, Florian and Krishnamoorthi, Raghuraman and Ghanem, Bernard and Chandra, Vikas and Xiong, Yunyang},
  journal={arXiv preprint arXiv:2601.05175},
  year={2026}
}

@article{feng2025onethinker,
  title={OneThinker: All-in-one Reasoning Model for Image and Video},
  author={Feng, Kaituo and Zhang, Manyuan and Li, Hongyu and Fan, Kaixuan and Chen, Shuang and Jiang, Yilei and Zheng, Dian and Sun, Peiwen and Zhang, Yiyuan and Sun, Haoze and others},
  journal={arXiv preprint arXiv:2512.03043},
  year={2025}
}

@article{feng2025video,
  title={Video-R1: Reinforcing Video Reasoning in MLLMs},
  author={Feng, Kaituo and Gong, Kaixiong and Li, Bohao and Guo, Zonghao and Wang, Yibing and Peng, Tianshuo and Wang, Benyou and Yue, Xiangyu},
  journal={arXiv preprint arXiv:2503.21776},
  year={2025}
}

@article{wang2025internvl3_5,
  title={InternVL3.5: Advancing Open-Source Multimodal Models in Versatility, Reasoning, and Efficiency},
  author={Wang, Weiyun and Gao, Zhangwei and Gu, Lixin and Pu, Hengjun and Cui, Long and Wei, Xingguang and Liu, Zhaoyang and Jing, Linglin and Ye, Shenglong and Shao, Jie and others},
  journal={arXiv preprint arXiv:2508.18265},
  year={2025}
}

@article{Qwen3-VL,
      title={Qwen3-VL Technical Report}, 
      author={Shuai Bai and Yuxuan Cai and Ruizhe Chen and Keqin Chen and Xionghui Chen and Zesen Cheng and Lianghao Deng and Wei Ding and Chang Gao and Chunjiang Ge and Wenbin Ge and Zhifang Guo and Qidong Huang and Jie Huang and Fei Huang and Binyuan Hui and Shutong Jiang and Zhaohai Li and Mingsheng Li and Mei Li and Kaixin Li and Zicheng Lin and Junyang Lin and Xuejing Liu and Jiawei Liu and Chenglong Liu and Yang Liu and Dayiheng Liu and Shixuan Liu and Dunjie Lu and Ruilin Luo and Chenxu Lv and Rui Men and Lingchen Meng and Xuancheng Ren and Xingzhang Ren and Sibo Song and Yuchong Sun and Jun Tang and Jianhong Tu and Jianqiang Wan and Peng Wang and Pengfei Wang and Qiuyue Wang and Yuxuan Wang and Tianbao Xie and Yiheng Xu and Haiyang Xu and Jin Xu and Zhibo Yang and Mingkun Yang and Jianxin Yang and An Yang and Bowen Yu and Fei Zhang and Hang Zhang and Xi Zhang and Bo Zheng and Humen Zhong and Jingren Zhou and Fan Zhou and Jing Zhou and Yuanzhi Zhu and Ke Zhu},
	  journal={arXiv preprint arXiv:2511.21631},
      year={2025}
}

@inproceedings{zhang2025mindpower,
  title={MindPower: Enabling Theory-of-Mind Reasoning in VLM-based Embodied Agents},
  author={Zhang, Ruoxuan and Zheng, Qiyun and Zhou, Zhiyu and Liao, Ziqi and Wu, Siyu and Jiang-Lin, Jian-Yu and Wen, Bin and Xie, Hongxia and Fu, Jianlong and Cheng, Wen-Huang},
  booktitle={Proceedings of the IEEE/CVF Conference on Computer Vision and Pattern Recognition (CVPR)},
  year={2026}
}

@misc{li2026roboclaw,
  title={RoboClaw: An Agentic Framework for Scalable Long-Horizon Robotic Tasks},
  author={Ruiying Li and Yunlang Zhou and YuYao Zhu and Kylin Chen and Jingyuan Wang and Sukai Wang and Kongtao Hu and Minhui Yu and Bowen Jiang and Zhan Su and Jiayao Ma and Xin He and Yongjian Shen and Yangyang and Guanghui Ren and Maoqing Yao and Wenhao Wang and Yao Mu},
  year={2026},
  eprint={2603.11558},
  archivePrefix={arXiv},
  primaryClass={cs.RO},
  url={https://arxiv.org/abs/2603.11558}, 
}

@misc{huo2026abotclawfoundationpersistentcooperative,
      title={ABot-Claw: A Foundation for Persistent, Cooperative, and Self-Evolving Robotic Agents}, 
      author={Dongjie Huo and Haoyun Liu and Guoqing Liu and Dekang Qi and Zhiming Sun and Maoguo Gao and Jianxin He and Yandan Yang and Xinyuan Chang and Feng Xiong and Xing Wei and Zhiheng Ma and Mu Xu},
      year={2026},
      eprint={2604.10096},
      archivePrefix={arXiv},
      primaryClass={cs.CV},
      url={https://arxiv.org/abs/2604.10096}, 
}

@misc{openclaw2026, title = {OpenClaw: Personal AI Assistant}, author = {{OpenClaw Contributors}}, year = {2026}, howpublished = {\url{https://github.com/openclaw/openclaw}}, note = {Open-source software} }

\end{document}